\documentclass[11pt]{article}
\usepackage{geometry}
\usepackage{coling2020}
\usepackage{times}
\usepackage{url}
\usepackage{latexsym}
\usepackage{microtype}
\usepackage{booktabs}

\usepackage[pdftex]{graphicx}
\graphicspath{{images/}}

\colingfinalcopy 

\title{aschern at SemEval-2020 Task 11:\\
It Takes Three to Tango: RoBERTa, CRF, and Transfer Learning}

\author{Anton Chernyavskiy\textsuperscript{1}, Dmitry Ilvovsky\textsuperscript{1} \and Preslav Nakov\textsuperscript{2} \\
\textsuperscript{1}National Research University Higher School of Economics, Moscow, Russia \\
  \textsuperscript{2}Qatar Computing Research Institute, HBKU, Doha, Qatar\\
  {\tt aschernyavskiy\_1@edu.hse.ru, dilvovsky@hse.ru} \\ 
  {\tt pnakov@hbku.edu.qa} \\}

\date{}

\begin{document}
\maketitle
\begin{abstract}
  We describe our system for SemEval-2020 Task 11 on \textit{Detection of Propaganda Techniques in News Articles}. We developed ensemble models using RoBERTa-based neural architectures, additional CRF layers, transfer learning between the two subtasks, and advanced post-processing to handle the multi-label nature of the task, the consistency between nested spans, repetitions, and labels from similar spans in training. We achieved sizable improvements over baseline fine-tuned RoBERTa models, and the official evaluation ranked our system $\mathbf{3^{rd}}$ (almost tied with the 2nd) out of 36 teams on the span identification subtask with an F1 score of 0.491, and $\mathbf{2^{nd}}$ (almost tied with the 1st) out of 31 teams on the technique classification subtask with an F1 score of 0.62.
\end{abstract}

\section{Introduction}

The proliferation of disinformation online, commonly known as ``fake news'', has given rise to a lot of research on automatic fake news detection. However, most of the efforts have focused on checking whether a piece of information is factually correct, and little attention has been paid to the propaganda techniques that malicious actors use to spread their message. 
SemEval-2020 Task 11 \cite{DaSanMartinoSemeval20task11} aims to bridge this gap. It focused on detecting the use of propaganda techniques in news articles,\footnote{The official task webpage: \url{http://propaganda.qcri.org/semeval2020-task11/}} creating a dataset that extends \cite{da-san-martino-etal-2019-fine}, and offering two subtasks:
\begin{itemize}
\item \textbf{span identification (SI):} detecting the propaganda spans in an article;
\item \textbf{technique classification (TC):} detecting the type of propaganda used in a given text span.
\end{itemize}

Below, we describe the systems we built for these two subtasks. At the core of our systems is RoBERTa \cite{DBLP:journals/corr/abs-1907-11692}, a pre-trained model based on the Transformer architecture \cite{NIPS2017_7181}. However, we improved over RoBERTa by adding extra layers in the neural network architecture, and we further added some post-processing steps. We further applied transfer learning between the two subtasks, and finally, we combined different models into an ensemble.\footnote{The code of our systems is available at \url{http://github.com/aschern/semeval2020_task11}}


\section{Related Work}
\label{sec:2}

The dataset for the task comes from \cite{da-san-martino-etal-2019-fine}, which used a BERT-based model with multi-task learning and a gated architecture; the system can be tried online~\cite{da-san-martino-etal-2020-prta}.
There was also a related previous task on fine-grained propaganda detection  \cite{da-san-martino-etal-2019-findings}, where the participants used Transformer-style models, LSTMs and ensembles \cite{fadel-etal-2019-pretrained,hou-chen-2019-caunlp,hua-2019-understanding}. Some approaches further used non-contextualized word embeddings, e.g.,~based on FastText and GloVe \cite{gupta-etal-2019-neural,al-omari-etal-2019-justdeep}, or handcrafted features such as LIWC, quotes and questions \cite{alhindi-etal-2019-fine}.  For the fragment-classification subtask (a combination of two subtasks of the current SemEval) the LSTM-CRF \cite{gupta-etal-2019-neural} or biLSTM-CRF \cite{alhindi-etal-2019-fine} models were applied besides BERT~\cite{propaganda:NeurIPS:2019}. Moreover, some efforts have been made to increase the size of the dataset using unsupervised language model pre-training \cite{yoosuf-yang-2019-fine}.

Finally, there is a recent survey on computational propaganda detection~\cite{IJCAI2020:propaganda:survey}.

\section{Our Systems}\label{sec:3}

In this section, we provide a general overview of our systems for the two subtasks. For both subtasks, we trained ensembles based on RoBERTa with some postprocessing.

\subsection{Subtask 1: Span Identification}

\paragraph{Model} We addressed the span identification subtask as a sequence labeling problem. To that end, we transformed the initial span markup into a BIO tagging format (Begin, Inside, Outside), as our preliminary experiments had shown that it performed better than alternatives such as IO and BIOUL (Begin, Inside, Outside, Unit, Last)~\cite{ratinov-roth-2009-design}. As we have only one possible entity class \texttt{PROP}, each token can be assigned one of the following three labels: \texttt{O}, \texttt{B-PROP}, and \texttt{I-PROP}. Then, we fine-tuned a RoBERTa model to predict the above BIO tags for each token in the input sentence. 

One problem with the above setup is that each token is classified independently of the surrounding tokens: while these surrounding tokens are taken into account in the contextualized embeddings that RoBERTa produces, there is no modeling of the dependency between the predicted labels: for example, logically \texttt{I-PROP} cannot follow \texttt{O}, but RoBERTa does not model it. Thus, we further added a linear-chain Conditional Random Field (CRF) model \cite{10.5555/645530.655813} as an additional layer, in order to model the dependency between the labels predicted for the individual tokens; this model can observe that the sequence \texttt{O I-PROP} is never present in the training data, and thus it can assign a very low probability to the transition from an \texttt{O} tag to an \texttt{I-PROP} tag. 

\begin{figure}[!t]
\centering
\includegraphics[scale=0.8]{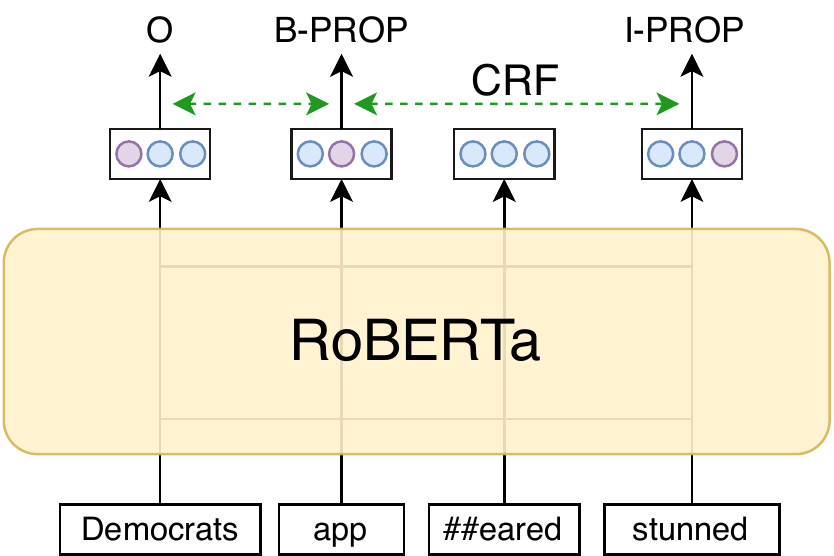}
\caption{\textbf{SI subtask:} Our RoBERTa-CRF model with BIO encoding. It is trained end-to-end and the CRF model ignores non-starting word pieces (\textit{\#\#eared} in the example).}
\label{fig_1}
\end{figure}

We trained the resulting RoBERTa-CRF model in an end-to-end fashion as shown in Figure~\ref{fig_1}. The CRF receives the logits for each input token, and makes a prediction for the entire input sequence, taking into account the dependencies between the labels, similarly to \cite{lample-etal-2016-neural}. Note that RoBERTa works with byte pair encoding (BPE) units, while for the CRF it makes more sense to work with words. Thus, in the input to the CRF, we only used tokens that started a word, and we skipped any word continuation tokens, e.g.,~a token like \textit{\#\#smth} would not be passed to the CRF.

\paragraph{Post-processing} We further applied two post-processing steps to obtain the final prediction from the token classification. First, we made sure that each predicted propaganda span began and ended by either a letter or a number (alphanumerical); otherwise, we shortened the span by advancing its beginning and/or by pushing back its ending by 1-2 characters until both the beginning and the ending characters became alphanumerical, which effectively defends against tokenization errors. Second, we checked if the span was preceded and followed by quotation marks, in which case we expanded it to include them; this is helpful as sometimes propaganda techniques contain text in quotations.

\paragraph{Ensemble} Finally, in order to increase the stability of the model, we created an ensemble of two models that have the same architecture, but are trained using different random seeds. At test time, each classifier made an independent prediction, and then we took the union of the predicted spans. In case of overlaps between spans, we created a superspan spanning the union of the respective overlapping spans.

\subsection{Subtask 2: Technique Classification}

\paragraph{Model} The technique classification subtask is a multi-class \emph{multi-label} problem, as it asks to predict one or more labels per span. As only a small number of examples in the training dataset have multiple labels (propaganda techniques), we reduced the problem to a multi-class \emph{single-label} problem. In particular, at training time, we converted all multi-label training examples into single-label ones by creating multiple copies of each multi-label example, one copy for each of the labels; at testing time, we considered the $n$-best predicted labels and we decided whether to predict more than one label in a post-processing step.

Our model for the TC subtask is based on RoBERTa, and it takes the following input: \texttt{[CLS] <span> [SEP] <sentence>}, where \texttt{<sentence>} is the sentence from which the span was extracted. 
Then, we added a softmax layer on top of the embedding for the \texttt{[CLS]} token to make a classification prediction, and we fune-tuned the model using the training data.

\begin{figure}[!t]
\centering
\includegraphics[scale=0.8]{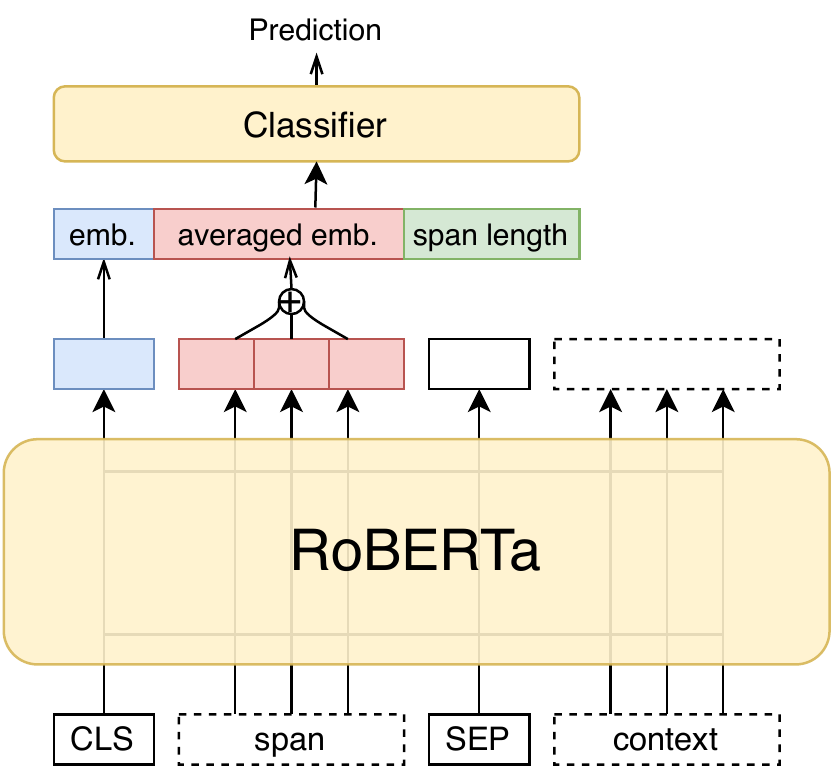}
\caption{\textbf{TC subtask:} Our RoBERTa-based model takes as an input a target span and its context (the sentence where the span originated). The classifier takes as an input the embedding of the [\textsc{CLS}] token, the averaged embedding of all span tokens, and the span length.}
\label{fig_2}
\end{figure}

We further developed a variant of the model, which concatenates (\emph{i})~the RoBERTa embedding for the \texttt{[CLS]} token, (\emph{ii})~the averaged embedding of the remaining tokens, and (\emph{iii})~the length of the span. Length is important, as some propaganda techniques such as \emph{Loaded Language} and \emph{Name Calling} are typically short, while others such as \emph{Causal Oversimplification} are generally longer.
Finally, we added an extra fully-connected layer with the size of the original RoBERTa embeddings, as shown on Figure~\ref{fig_2}. 

We further used transfer learning from the span identification subtask: we first trained the model using the data for the span identification subtask, and then we continued training for the technique classification subtask. As a result, the embeddings model how propagandistic these tokens are, which in turn can help discriminate between different types of propaganda for the TC subtask, as some propanganda techniques, such as \emph{Loaded Language} and \emph{Name Calling}, are short and their tokens are likely to be highly propagandistic, while other propaganda techniques are long, such as \emph{Red Herring} and \emph{Causal Oversimplification}, and many of the tokens in their spans are not propagadistic by themselves.

\paragraph{Post-processing} While developing our system, we noticed that our model struggled with \emph{Repetition}, as repetitions can occur over long distances that go beyond the maximum span length that RoBERTa can handle: 512 tokens. Thus, we added some special post-processing to handle this technique. Since all candidate spans are given for the TC subtask, we compared these spans looking for possible repetitions. We compared the spans looking for exact matches after removing punctuation, filtering out stopwords, and applying the Porter stemmer~\cite{porter1980algorithm}.
We assigned a \emph{Repetition} label in case the span matched at least two other spans. If it matched only one other span, we further required that the classifier predicted \emph{Repetition} with a probability greater than 0.001. If it matched no other spans, we assigned \emph{Repetition} a probability of 0.0, unless the classifier had predicted \emph{Repetition} with probability of 0.99 or higher.

We further checked for each test span whether it can be found as a span in the training data (as above, for the matching, we ignored punctuation, stopwords, and we used stemming), and if so, we first collected all the propaganda techniques that the span has been seen with in the training data (note that the span might occur multiple times in training, possibly with different annotations for the different instances, and some instances could have multiple techniques assigned), and then we boosted the corresponding predicted probabilities by 0.5. This works well for spans that are likely to express the same propaganda technique(s) regardless of the context.

Next, we modeled the local consistency of the predicted spans. We observed that some long propaganda spans could contain a subspan with a different propaganda technique. We further noticed that not all combinations were equally likely, e.g.,~\emph{Causal Oversimplification} was generally long and it could contain a subspan of \emph{Loaded Language}, but it was very unlikely to see these propaganda techniques nested the other way around. Thus, we collected all possible span-subspan combinations of propaganda techniques observed in training, and we tried to discourage any other combinations. For any span-subspan combination that was not observed in training, we assigned 0.0 to the smallest of the two probabilities for the considered spans, unless the new maximum probability for the affected span would drop more than twice as a result.

Moreover, we modeled the multi-label nature of the TC subtask. We took advantage of the fact that in case a given span had to be assigned $n$ propaganda techniques ($2<=n<=14$), this span would be repeated $n$ times in the test input, i.e.,~the number $n$ was known, and only the actual propaganda techniques were to be predicted. We handled this by assigning the top-$n$ propaganda techniques for such a span, according to the calculated probabilities (after they have been potentially altered by the previous post-processing steps).

\paragraph{Ensemble} Finally, we used model combination. In particular, we combined the simple RoBERTa model with the more sophisticated one from Figure~\ref{fig_2}: we took the posterior probabilities they produced for all propaganda techniques, and we passed them to a logistic regression classifier to make the final decision. We tuned the parameters of the classifier on part of the development dataset.

\section{Experimental Setup}\label{sec:4}

\paragraph{Data} We experimented with the training, the development and the test datasets provided for SemEval-2020 Task 11, which contain 371, 75, and 90 news articles with 6,128, 1,063 and 1,790 spans, respectively. 
We randomly selected 20\% of the training dataset for local evaluation when developing our models.

\paragraph{Evaluation measures} The official evaluation measures are a ``normalized'' version of the F1 score for the span identification subtask, and micro-averaged F1 score for the technique classification subtask.
A detailed description of the evaluation measures can be found in the SemEval-2020 task 11 paper \cite{DaSanMartinoSemeval20task11}.

\paragraph{Parameter settings} We used the RoBERTa-large model and the following hyper-parameters, which we selected using validation on a subsample of the training data: learning rate of 2e-5, batch size of 24, and RoBERTa's default optimizer with 500 warm-up steps. We further found that an uncased model should be used for the span identification subtask, but that a cased model worked better for the technique classification subtask. We trained the models for 30 epochs, we saved a checkpoint every two epochs, and we selected the best checkpoint on the validation subsample.

\section{Results}\label{sec:5}

\subsection{Subtask 1: Span Identification}

On the development set, our fine-tuned BIO-encoded RoBERTa-large model achieved an F1 score of 47.8; see Table~\ref{SI-analysis-table}. Adding a CRF layer pushed F1 to 48.8, and also yielded higher stability of the results, i.e.,~less variation across reruns. The ensemble of these two models improved F1 score to 49.6. Finally, adding post-processing of punctuation and quotation symbols pushed the final F1 score to 49.9.

\begin{table*}[t!]
\begin{center}
\begin{tabular}{lc}
\toprule
\bf System & \bf F1 \\
\midrule
RoBERTa-large with BIO encoding (fine-tuned) & 0.478  \\ 
\hspace{10pt}+ CRF layer & 0.488 \\ 
\hspace{26pt}+ ensemble & 0.496 \\ 
\hspace{42pt}+ post-processing & 0.499 \\
\bottomrule
\end{tabular}
\end{center}
\caption{\label{SI-analysis-table} \textbf{SI subtask (dev):} An incremental analysis of our system.}
\end{table*}

\begin{table}[t!]
\begin{center}
\begin{tabular}{clrrr}
\toprule
\bf Rank & \bf Team & \bf F1  & \bf Prec.  & \bf Recall \\
\midrule
1 & Hitachi & 51.55 & 56.54  & 47.37 \\
2 & ApplicaAI & 49.15  & 59.95  & 41.65 \\
3 & \textbf{aschern} & 49.10  & 53.23 & 45.56 \\
4 & LTIatCMU & 47.66  & 50.97  & 44.76 \\
\midrule
36 & baseline & 0.31 & 13.04 & 0.16\\
\bottomrule 
\end{tabular}
\end{center}
\caption{\label{SI-test-table} \textbf{SI subtask (test):} Top-4 teams on the test set.}
\end{table}

The official results on the blind test dataset are shown in Table \ref{SI-test-table}. We can see that our official submission is ranked third and is almost tied with the second one, with a difference of only 0.05 F1 points absolute.

\subsection{Subtask 2: Technique Classification}

On the development set, our fine-tuned  RoBERTa-large model achieved an F1 score of 62.18; see Table~\ref{TC-analysis-table}. Adding length and averaged span embeddings, improved the results only marginally. Marginal was also the improvement from multi-label correction and from bonus for span labels seen on training, which together only added 0.5 F1 points absolute. However, handling of \emph{Repetition} yielded a huge improvement of over 3.5 F1 points absolute. Further adding checking for unseen span-subspan combinations yielded marginal gains. Finally, using an ensemble improved the F1 score by 1.5 F1 points absolute to 68.10.

The official results on the blind test dataset are shown in Table \ref{TC-test-table}. We can see that our official submission is ranked second and is almost tied with the first team, with a difference of only 0.06 F1 points absolute.

\section{Conclusion and Future Work}
\label{sec:6}

We described the system we developed for the SemEval-2020 Task 11 on Detection of Propaganda Techniques in News Articles. We developed ensembles of RoBERTa-based neural architectures, additional CRF layers, transfer learning between the two subtasks, and advanced post-processing to handle the multi-label nature of the task, the consistency between nested spans, repetitions, and labels from similar spans in training. We achieved sizable improvements over baseline RoBERTa models, and the official evaluation ranked our system $\mathbf{3^{rd}}$ (almost tied with the 2nd) out of 36 teams on the span identification subtask, and $\mathbf{2^{nd}}$ (almost tied with the 1st) out of 31 teams on the techniques classification subtask.

In future work, we plan to explore other neural architectures such as T5 \cite{raffel2019exploring} and GPT-3 \cite{brown2020language}. We further want to explore transfer learning from other tasks such as argumentation mining \cite{argumentation:mining:book} and offensive language detection \cite{OffenseEval:NAACL:2019,zampieri-etal-2020-semeval}.

\begin{table*}[t!]
\begin{center}
\begin{tabular}{lc}
\toprule
\bf System & \bf F1 \\
\midrule
RoBERTa-large (fine-tuned) & 62.18  \\ 
\hspace{10pt}+ length + averaged span embeddings & 62.37 \\ 
\hspace{26pt}+ post-processing: multi-label correction & 62.75 \\ 
\hspace{42pt}+ post-processing: bonus for span labels seen on training & 62.85 \\ 
\hspace{58pt}+ post-processing: handling of \emph{Repetition} & 66.42 \\
\hspace{74pt}+ post-processing: checking for unseen span-subspan combinations & 66.60 \\ 
\hspace{90pt}+ ensemble & \textbf{68.10} \\
\bottomrule
\end{tabular}
\end{center}
\caption{\label{TC-analysis-table} \textbf{TC subtask (dev):} An incremental analysis of our system on the development set.}
\end{table*}

\begin{table}[t!]
\begin{center}
\begin{tabular}{clc}
\toprule
\bf Rank & \bf Team & \bf F1\\
\midrule
1 & ApplicaAI & 62.07 \\
2 & \textbf{aschern} & 62.01 \\
3 & Hitachi & 61.73 \\
4 & Solomon & 58.94 \\
\midrule
30 & baseline & 25.20 \\
\bottomrule
\end{tabular}
\end{center}
\caption{\label{TC-test-table} \textbf{TC subtask (test):} Results on the test set (top-4 teams).}
\end{table}

\section{Acknowledgments}

Anton Chernyavskiy and Dmitry Ilvovsky performed this research within the framework of the HSE University Basic Research Program, funded by the Russian Academic Excellence Project `5-100'.

Preslav Nakov contributed as part of the Propaganda Analysis
Project (\url{propaganda.qcri.org}), part of the Tanbih megaproject (\url{tanbih.qcri.org}), developed at the Qatar Computing Research Institute, HBKU. Tanbih aims to limit the effect of ``fake 
news'', propaganda, and media bias by making users aware of what they are reading, thus promoting media literacy and critical thinking. 

\bibliographystyle{coling}
\bibliography{semeval2020}

\end{document}